\documentclass[conference]{IEEEtran}
\IEEEoverridecommandlockouts
\usepackage{cite}
\usepackage[utf8]{inputenc}
\usepackage{comment}
\usepackage{array}
\usepackage{longtable}
\usepackage{xspace} 
\usepackage{color}
\usepackage{multirow}
\usepackage{amsmath}
\usepackage{mathtools}
\usepackage{graphicx}
\usepackage{tabularx}
\usepackage{verbatim}
\usepackage{xcolor}
\usepackage{color, colortbl}

\newcolumntype{Y}{>{\centering\arraybackslash}X}
\newcolumntype{L}{>{\raggedleft\arraybackslash}X}

\definecolor{10}{rgb}{0,0,0} 
\definecolor{89}{rgb}{0.2,0.2,0.2}
\definecolor{67}{rgb}{0.35,0.35,0.35}
\definecolor{45}{rgb}{0.5,0.5,0.5}
\definecolor{23}{rgb}{0.65,0.65,0.65}
\definecolor{01}{rgb}{0.8,0.8,0.8}

\usepackage{enumitem,kantlipsum}
\usepackage{threeparttable}
\usepackage{color, colortbl}
\newcolumntype{Y}{>{\centering\arraybackslash}X}
\newcolumntype{L}{>{\raggedleft\arraybackslash}X}

\definecolor{RRRR}{rgb}{0,0,0} 
\definecolor{RRRO}{rgb}{0.1,0.1,0.1} 
\definecolor{RROO}{rgb}{0.2,0.2,0.2}
\definecolor{ROOO}{rgb}{0.3,0.3,0.3}
\definecolor{OOOO}{rgb}{0.4,0.4,0.4}
\definecolor{OOOY}{rgb}{0.5,0.5,0.5}
\definecolor{OOYY}{rgb}{0.6,0.6,0.6}
\definecolor{OYYY}{rgb}{0.7,0.7,0.7}
\definecolor{YYYY}{rgb}{0.8,0.8,0.8}

\def\BibTeX{{\rm B\kern-.05em{\sc i\kern-.025em b}\kern-.08em
    T\kern-.1667em\lower.7ex\hbox{E}\kern-.125emX}}

\begin{document}
\title{Promoting Semantics in Multi-objective Genetic Programming based on Decomposition}

\author{
\IEEEauthorblockN{Edgar Galv\'an$^*$\thanks{$^*$Main and senior author.}}
\IEEEauthorblockA{{Naturally Inspired Computation Research Group} \\
{Department of Computer Science},
{Hamilton Institute}\\
Maynooth University, Ireland \\
edgar.galvan@mu.ie}
\and
\IEEEauthorblockN{Fergal Stapleton}
\IEEEauthorblockA{{Naturally Inspired Computation Research Group} \\
{Department of Computer Science},
{Hamilton Institute}\\
Maynooth University, Ireland \\
fergal.stapleton.2020@mumail.ie}

}

\IEEEoverridecommandlockouts
\IEEEpubid{\makebox[\columnwidth]
{} 
\hspace{\columnsep}\makebox[\columnwidth]{ }}

\maketitle
\IEEEpubidadjcol

\begin{abstract}
  The study of semantics in Genetic Program (GP) deals with the behaviour of a program given a set of inputs and has been widely reported in helping to promote diversity in GP for a range of complex problems ultimately improving evolutionary search. The vast majority of these studies have focused their attention in single-objective GP, with just a few exceptions where Pareto-based dominance algorithms such as NSGA-II and SPEA2 have been used as frameworks to test whether highly popular  semantics-based methods, such as Semantic Similarity-based Crossover (SSC), helps or hinders evolutionary search. Surprisingly it has been reported that the benefits exhibited by SSC in SOGP are not seen in Pareto-based dominance Multi-objective GP. In this work, we are interested in studying if the same carries out in Multi-objective Evolutionary Algorithms based on Decomposition (MOEA/D). By using the MNIST dataset, a well-known dataset used in the machine learning community, we show how SSC in MOEA/D promotes semantic diversity yielding better results compared to when this is not present in canonical MOEA/D.

\end{abstract}


\section{Introduction}

Semantics in  Genetic Program (GP)~\cite{koza_1994_genetic} can be understood as the behaviour of a program given a (partial) set of inputs and has been well documented in improving performance of GP~\cite{Galvan-Lopez2016,DBLP:conf/ppsn/MoraglioKJ12,Vanneschi2013,Uy2011}. There have been a wide variety of different methods which have incorporated semantics into single-objective GP (SOGP) but thus far research into implementing semantics into Multi-objective GP (MOGP) has limited and centred around only two algorithms~\cite{DBLP:conf/gecco/GalvanS19,Galvan-Lopez2016,Galvan_MICAI_2016,DBLP:journals/corr/abs-2009-12401}: 
the Non-dominated Sorting Genetic Algorithm \sloppy{(NSGA-II)~\cite{Deb02afast}} and the Strength Pareto Evolutionary Algorithm  (SPEA2)~\cite{Deb02afast,Zitzler01spea2:improving}. Both methods utilize the concept of Pareto-dominance in the fitness criteria for retaining individuals into subsequent generations. A candidate solution is said to be Pareto dominant if its fitness is better or equal for all objectives and is strictly preferred by at least one in the search space.

Semantic similarity-based crossover (SSC) was originally proposed by Uy et al.~\cite{Uy2011} in the context of single-optimization GP. This method uses a computationally (potentially expensive) procedure by applying crossover between two parents multiple times using semantic diversity as a criteria in the crossover process and has proved to be highly beneficial in multiple independent research studies where SSC has been used in SOGP~\cite{DBLP:conf/eurogp/AnhNNO13,6557931,Uy2011,10.1007/978-3-642-04962-0_7}. However, it has been reported in~\cite{DBLP:conf/gecco/GalvanS19,Galvan-Lopez2016,Galvan_MICAI_2016} that SSC does not have the same positive impact when using the concept of Pareto-dominance, as demonstrated by the authors using NSGA-II and SPEA2 in their empirical studies.

The goal of this paper is to demonstrate whether SSC has a positive impact or not in MOGP by using decomposition instead of Pareto-dominance, where it has demonstrated before that SSC does not help evolutionary search when using the latter. To the best of our knowledge this is the first study where the highly popular form of SSC used in SOGP is studied in the context of Multi-objective Evolutionary Algorithm (GP in this case) based on decomposition.

The outline of the paper is a follows; Section \ref{sec:rel} deals with related work covering semantics in GP, MOGP in general as well as discussing work relating to MOEA/D, Section \ref{sec:method} deals with method proposed in this study. The experimental setup is discussed in Section~\ref{sec:sub:experiment} and Section~\ref{sec:results} discusses our results and analysis thereof and is followed by our concluding remarks as well as a brief discussion on future work in Section~\ref{sec:conclusions}.

\section{Related Work}
\label{sec:rel}

GP is a form of Evolutionary Algorithm (EA) that uses genetic operations that are analogous to behavioural biology and evolve programs towards finding a solution to a problem. The range of problem domains for GP are wide and this form of EA has been found to be beneficial for problems with multiple local optima and for problems with a varying degree of complexity~\cite{eiben_2015_from}, making EAs ideal for highly complex problems including the automatic configuration of deep neural networks' architectures and their training (an in-depth recent literature review in this emerging research area can be found in~\cite{galvan2020neuroevolution}). However despite the well documented effectiveness of canonical GP, there are well-known limitations of these methods, through the study of properties of encodings~\cite{DBLP:conf/cec/LopezMOB10,Galvan-Lopez2011}, and research is on going into finding and developing approaches to improve their overall performance including promoting neutrality in deceptive and challenging landscapes~\cite{DBLP:conf/eurogp/LopezDP08,DBLP:conf/ppsn/LopezP06_2,10.1007/978-3-540-73482-6_9,DBLP:journals/tec/PoliL12}, dynamic fitness cases~\cite{DBLP:conf/gecco/LopezVST17Poster,DBLP:conf/ae/LopezVST17}, reuse of code~\cite{DBLP:conf/eurogp/LopezPC04}, variants of GP~\cite{Galvan-Lopez2008}, use of surrogate models~\cite{DBLP:conf/cec/KattanOL13,6256108} to mention a few examples. Successful examples of these improvements can be found in applicable areas including energy-based problems~\cite{galvan_neurocomputing_2015,DBLP:conf/ijcci/LopezSP15}.


An area of research which has proven popular in advancing the field of GP has been semantics. There have been a number of definitions for semantics in the past but broadly speaking semantics can be defined as the behavioural output of a program when executed using a set of inputs.

\subsection{Semantics in Genetic Programming}

Scientific studies of semantics in GP have increased dramatically over the last years given that it has been consistently reported to be beneficial in GP search, ranging from the study of geometric operators~\cite{DBLP:conf/ppsn/MoraglioKJ12}, including the analysis of indirect semantics~\cite{Galvan-Lopez2016,Uy2011} up to the use of semantics in real-world problems~\cite{Vanneschi2013}.  We discuss next the most relevant works to the research discussed in this paper. 

Even though researchers have proposed a variety of \sloppy{mechanisms} to use the semantics of GP programs to guide a search, it is commonly accepted that semantics refers to the output of a GP program once it is executed on a data set (also known as fitness cases in the specialized GP literature). The work conducted by McPhee et al.~\cite{McPhee:2008:SBB:1792694.1792707} paved the way for the proliferation of indirect semantics works. In their research, the authors studied the semantics of subtrees and the semantics of context (the remainder of a tree after the removal of a subtree). In their studies, the authors pointed out how a high proportion of individuals created by the widely used 90-10 crossover operator (i.e., 90\%-10\% internal-external node selection policy) are semantically equivalent. That is, the  crossover operator does not have any useful impact on the semantic GP space, which in consequence leads to a lack of performance increase as evolution continues. \par


With the goal to overcome the lack of semantic diversity reported by McPhee et al.~\cite{McPhee:2008:SBB:1792694.1792707}, Beadle and Johnson~\cite{4630784} proposed a Semantically Driven Crossover (SDC) operator that promotes semantic diversity over generations. More specifically, they used reduced ordered binary decision diagrams \sloppy{(ROBDD)} on Boolean problems (i.e., Multiplexer and the even-5-parity problem) to check for semantic similarity between parents and offspring. The authors showed a significant gain in terms of fitness improvement when promoting semantic diversity using the SDC operator. They also showed that by using ROBDD on these particular problems, the SDC operator was able to considerably reduce bloat (dramatic increase of tree sizes as evolution proceeds with a proportional improvement in fitness).  Later, Beadle and Johnson~\cite{4983099} also explored a similar technique for the mutation operator, called Semantically Driven Mutation (SDM). The authors inspected the use of semantics when applied to subtree mutation; i.e., this operator replaces the tree at any node with a randomly generated subtree.  The resulting individual is checked for semantic equivalence through reducing it to a canonical form which can be easily compared to parent trees that were similarly reduced. SDM increases bloat more than SDC, and when both operators are used, bloat is greatly increased. More recently, Fracasso and Von Zuben demonstrated new SDM operators that reduced bloat without suffering from performance loss.\par
These works, however, used discrete fitness-valued cases, impeding their findings to be generalized in continuous search spaces. Uy et al.~\cite{Nguyen:2009:SAC:1533497.1533524} addressed this limitation and multiple works followed their approach thanks to its simplicity.
Uy et al. proposed four different forms of applying semantic crossover operators on real-valued problems (e.g., symbolic regression problems). To this end, the authors measured the semantic equivalence of two expressions by measuring them against a random set of points sampled from the domain. If the resulting outputs of these two expressions were close to each other, subject to a threshold value called semantic sensitivity, these expressions were regarded as semantically equivalent. In their experimental design, the authors proposed four scenarios. In their first two scenarios, Uy et al. focused their attention on the semantics of subtrees. More specifically, for Scenario I, they tried to encourage semantic diversity by repeating crossover for a number of trials if two subtrees were semantically equivalent. Scenario II explored the opposite idea of Scenario I. For the last two scenarios, the authors focused their attention on the entire trees. That is, for Scenario III Uy et al. checked if offspring and parents were semantically equivalent. If so, the parents were transmitted into the following generation and the offspring were discarded. The authors explored the opposite of this idea in Scenario IV (children semantically different from their parents). They showed, for a number of symbolic regression problems, that Scenario I produced better results compared to the other three scenarios. \par
The major drawback with the Uy et al.~\cite{Nguyen:2009:SAC:1533497.1533524} approach is that it can be computational expensive, since it relies on a trial mechanism that attempts to find semantically different individuals via the execution of the crossover operator multiple times. To overcome this limitation, Galv\'an et al.~\cite{6557931} proposed a cost-effective mechanism based on the tournament selection operator to promote semantic diversity. More specifically, the tournament selection of the first parent is done as usual. That is, the fittest individual is chosen from a pool of individuals randomly picked from the population. The second parent is chosen from a pool of individuals that are semantically different from the first parent and it is also the fittest individual. If there is no individual semantically different from the first parent, then the tournament selection of the second parent is performed as usual. The proposed approach resulted in similar, and in some cases better, results compared to those reported by Uy et al.~\cite{Nguyen:2009:SAC:1533497.1533524,Uy2011} without the need of a trial and error (expensive) mechanism.


More recently, Forstenlechner et al.~\cite{Forstenlechner:2018:TES:3205455.3205592} investigated the use of semantics in a different domain from those discussed before (Symbolic Regression and Boolean problems). The authors showed how it is possible to compute the semantics of GP individuals for program synthesis.
This operates on a range of different data types as opposed to those working on a single type of data. They computed the semantics of a GP individual by tracing it, which shows the behaviour of a program. To promote semantic diversity the authors used two approaches to determine semantic similarity, named (1) `partial change', used in the first instance, and (2) `any change', used if the first instance fails to be satisfied, to try to avoid using standard crossover. 
`Partial change' is when at least one entry has to be different and when at least one entry has to be identical, for a pair of semantic results stored in vectors. On the other hand, `any change' does not have the constraint shown in `partial change'. In their results, the authors reported that a semantic-based GP system achieved better results in four out of eight problems used in their work.   


\subsection{Multi-objective Genetic Programming Based on Pareto-dominance}

In a multi-objective optimization (MO) problem, one optimizes with respect to multiple goals or objective functions. Thus, the task of the algorithm is to find acceptable solutions by considering all the criteria simultaneously. This can be achieved in various ways, where  keeping the objectives separate is the most common. This form keeps the objectives separate and uses the notion of \textit{Pareto dominance}. In this way, Evolutionary MO (EMO)~\cite{1597059,CoelloCoello1999,Deb:2001:MOU:559152} offers an elegant solution to the problem of optimizing two or more conflicting objectives. The aim of EMO is to simultaneously evolve a set of the best trade-off solutions along the objectives in a single run.  EMO is one of the most active research areas in EAs thanks to its wide applicability as well as the impressive results achieved by these techniques~\cite{1597059,CoelloCoello1999,Deb:2001:MOU:559152}. Multi-objective optimisation is widely used outside EAs e.g.,~\cite{DBLP:conf/ijcnn/TaylorDLCC14,taylor:2013}.


MOGP has been used, for instance, to classify highly unbalanced binary data~\cite{6198882,Galvan-Lopez2016}. To do so, the authors treated each objective (class) `separately' using EMO approaches~\cite{Deb02afast,Zitzler01spea2:improving}. Bhowan et al.~\cite{6198882} and Galv\'an et al.~\cite{DBLP:conf/gecco/GalvanS19,Galvan-Lopez2016,Galvan_MICAI_2016} showed, independently, how MOGP was able to achieve high accuracy in classifying binary data in conflicting learning objectives (i.e., normally a high accuracy of one class results in lower accuracy on the other).

\subsection{Multi-objective Evolutionary Algorithms based on Decomposition}

Decomposition methods `decompose' a multi-objective problem into a set of sub-problems. These sub-problems are optimized simultaneously using a finite set of neighbouring sub-problems and in doing so reduce the relevant computational complexity when compared to dominance-based methods such as NSGA-II~\cite{Deb02afast} and SPEA2~\cite{Zitzler01spea2:improving}. It has also been demonstrated that for some complex Pareto sets, Multi-objective Evolutionary Algorithms based on Decomposition (MOEA/D) outperformed NSGA-II~\cite{4633340}.

The most popular approaches to decomposition in MOEA/D have been the Weighted Sum, Tchebycheff and Penalty-based Boundary Intersection (PBI) approaches~\cite{4358754}. There have been also a number of variants to these methods that have produced better approximations for more complex Pareto fronts. Examples include MOEA/D Adaptive Weight Adjustment (AWA)~\cite{6818677} which uses adaptive weights to produce more uniform solutions and improve diversity and inverted PBI~\cite{10.1145/2576768.2598297} which is an extension of PBI and which also has performance benefits for more complex Pareto sets.
More recently,  Zheng et al.~\cite{Zheng2018} used a new decomposition approach that insured the distribution of weight vectors were more effective in promoting diversity and ensuring convergence of solutions in objective space. They found that properties of the Weighted Sum and Tchebycheff approaches, both had benefits and shortcomings and so developed a method that combined both these approaches into one.

\section{Method}
\label{sec:method}

Before presenting the methods used in this study, we lay the foundations of semantics. 

\subsection{Semantics}
\label{sec:sub:semantics}

Pawlak et al. \cite{6808504} gave a formal definition for program semantics. Let $p \in P$ be a program from a given programming language $P$. The program $p$ will produce a specific output $p(in)$ where input $in \in I$. The set of inputs $I$ can be understood as being mapped to the set of outputs $O$ which can be defined as $p:I \rightarrow O$. 

\textbf{ Def 1.} \emph{Semantic mapping function is a function $s:P \rightarrow S$ mapping any program $p$ from $P$ to its semantic $s(p)$, where we can show the semantic equivalence of two programs}. Eq~\ref{eq:def1} expresses this formally,

\begin{equation}
  s(p_1) = s(p_2) \iff  \forall\ in \in I: p_1(in) = p_2(in)
  \label{eq:def1}
\end{equation}

\noindent This definition presents three important and intuitive properties for semantics:

\begin{enumerate}
\item Every program has only one semantic attributed to it.
\item Two or more programs may have the same semantics.
\item Programs which produce different outputs have different semantics.
\end{enumerate}

\noindent In Def. 1, we have not given a formal representation of semantics. In the following, semantics will be represented as a vector of output values which are executed by the program under consideration using an input set of data. For this representation of semantics we need to define semantics under the assumption of a finite set of fitness cases, where a fitness case is a pair comprised of a program input and its respective program output $I$ $\times$ $O$. This allows us to define the semantics of a program as follows.

\textbf{Def 2.} \emph{The semantics $s(p)$ of a program $p$ is the vector of
values from the output set $O$ obtained by computing $p$ on all inputs from the input set $I$}. This is formally expressed in Eq.~\ref{eq:def2},

\begin{equation}
  s(p) = [p(in_1), p(in_2), ... , p (in_l)]
  \label{eq:def2}
\end{equation}

\noindent where $l = |I|$ is the size of the input set. 

\subsection{Multi-objective Evolutionary Algorithm with Decomposition}
\label{sec:sub:moead}
Multi-objective Evolutionary Algorithm based on Decomposition (MOEA/D) differs with Pareto-dominance based methods in that it decomposes the multi-objective problem into a subset of scalar optimisation problems~\cite{4358754}. It is important to note that with MOEA/D we still wish to approximate the Pareto-optimal front, but instead of using dominance to determine the fitness of our solutions we use a scalar value aggregated from multiple objectives method~\cite{4633340}. In the original seminal paper by Zhang et al.~\cite{4358754}, they proposed three scalarization methods:  Weighted sum, Tchebycheff and Penalty-based Boundary Intersection (PBI). As the weighted sum method was reported to poorly approximate concave fronts in their entirety and as the PBI method incorporates its own diversity preserving measure, our research focused primarily on the Tchebycheff approach, formally defined in Eq.~\ref{eq:tche},

\begin{equation}
  \min(g(x|\lambda)) = \underset{1 \leq j \leq m}{\max}  \{\lambda \dot | f_j(x) - z_{j}\}
  \label{eq:tche}
\end{equation}

\noindent where $\lambda$ is a weight vector that is assigned to each sub problem and represents a search direction in objective space, and $Z_j$ is the ideal point and represents the ideal solution for a given problem.

To compute this, the following steps are considered, \\

\noindent\textbf{Step 1: Initialization} \\

\noindent\textbf{Step 1.1)} Initialize external population $EP = \emptyset$. \\

\noindent\textbf{Step 1.2)} Calculate the Euclidean distance between any two weight vectors and find the $T$ closest weight vectors to each respective weight vector. For each $i = \{1, 2, \cdots,  N\}$, set $B(i) = \{i_1, i_2, \cdots, i_T\}$ where $B(i)$ can be understood as a neighbourhood reference table of indices and where $\lambda^{i_1}, \lambda^{i_2}, ... , \lambda^{i_T}$ are the $T$ closest weight vectors to $\lambda^{i}$.\\

\noindent\textbf{Step 1.3)} Randomly create the initial population $x^1, x^2 ... x^N$ and set the fitness value $FV^i = F(x^i)$.\\

\noindent\noindent\textbf{Step 2: Update} For $i = \{1, 2, \cdots,  N\}$; do the following steps\\

\noindent\textbf{Step 2.1)} Select two indices $k$ and $l$ randomly from the neighbourhood reference table $B(i)$ and generate new offspring $y$ from parents $x^k$ and $x^l$ by applying genetic operations.\\

\noindent\textbf{Step 2.2)} An optional problem-specific repair and improvement heuristic on $y$ to produce $y^\prime$, otherwise let y = $y^\prime$.\\

\noindent\textbf{Step 2.3)} Update $z$ such that for each j = $\{1, 2, \cdots , m\}$ if $z_j$ $<$ $f_j(y^\prime)$, then set $z_j = f_j(y^\prime)$. In the case where objective is to minimize F(x) then this inequality is reversed.\\

\noindent\textbf{Step 2.4)} Update the neighbouring solutions for the $i^{th}$ case such 

\subsection{Semantic Similarity-based Crossover MOGP}
\label{sec:sub:ssc}
To incorporate semantics in a MOGP paradigm, we first use the Semantic Similarity-based Crossover (SSC) originally proposed by Uy et al.~\cite{Uy2011} which, to the best of our knowledge, has been exclusively and successfully used in single-objective GP. Although some attempts have been made in~\cite{DBLP:conf/gecco/GalvanS19}, where the authors have indicated that the benefits of promoting diversity via SSC in Pareto-based dominance algorithms, such as NSGA-II and SPEA2, are limited, if any.

To use SSC in single-objective GP, a semantic distance must be computed first. Using Def. 2, this distance is obtained by computing the average of the absolute difference of values for every $in \in I$ between parent and offspring. If the distance value lies within a range, defined by one or two threshold values, then crossover is used to generate offspring. Because this condition may be hard to satisfy, the authors tried to encourage semantic diversity by repeatedly applying crossover up to 20 times. If after this, the condition is not satisfied, then crossover is executed as usual.

In this work, we decide to use a single threshold value to compute the semantic distance due to its simplicity as well as it has been proved that one threshold is enough to successfully promote semantic diversity~\cite{DBLP:conf/gecco/GalvanS19,Galvan-Lopez2016,Nguyen2016}. We can compute the semantic distance between parent $p$ and offspring $v$ using the  distance formally defined in Eq. (\ref{semantic:distance:one:value}),

\begin{equation}
\label{semantic:distance:one:value}
  d(p_j,v) = \sum_{i=1}^l 1 \text{ if }   |p(in_i) - v(in_i)| > \text{UBSS}  
\end{equation}


SSC made a notable impact in GP, showing, for the first time, how semantic diversity can be promoted in continuous search spaces, with several subsequent
papers following along this line~\cite{DBLP:conf/eurogp/AnhNNO13,DBLP:conf/gecco/GalvanS19,6557931,Krawiec:GPEM2013,6063448}. SSC is incorporated with the MOEA/D method as discussed previously.


\section{Experimental setup}
\label{sec:sub:experiment}
MNIST~\cite{726791} is a highly popular data set of hand written digits where the goal is to classify each of the 10 digits, from 0 to 9,  correctly. In the following experiments we have taken part of the MNIST where we took a subset of the data with 6000 entries for each digit. Each digit is considered in isolation, where the digit under consideration was classified as 1 and all other digits were classified as 0. As a result, the data itself is imbalanced with a ratio of 1 to 9. When splitting the training and test set, this imbalance ratio was maintained for each set. 
Feature extraction was performed by splitting each image into a series of boxes and returning the mean and standard deviation for each box.

The MNIST data set was split 50/50 with half of the entries being attributed to the training set and the other half for the test set. All the results reported in this work are based on the latter. The same class imbalance ratio is kept between the training and test set. 
The function set is the list of arithmetic operators used by the GP and are assigned at the non-terminal nodes, these were selected as $ \Re = \{+, -, *, \%, IF\} $, where \% denotes the protected division operator. The terminal set is defined as the set of problem features from the MNIST data set. These are the 18 features representing the mean and standard deviation of each image cell.

\begin{table}
\centering
\caption{Summary of parameters}
\resizebox{0.75\columnwidth}{!}{ 
\small\begin{tabular}{|l|r|} \hline 
\emph{Parameter} &
\emph{Value} \\ \hline \hline
Population Size & 500 \\ \hline
Generations & 50 \\ \hline
Type of Crossover & 90\% internal nodes, 10\% leaves  \\ \hline
Crossover Rate  & 0.60  \\ \hline
Type of Mutation & Subtree \\ \hline
Mutation Rate & 0.40 \\ \hline
Selection & Tournament (size = 7) \\ \hline
Initialisation Method & Ramped half-and-half \\ \hline
Initialisation Depths: & \\ 
\hspace{.3cm}Initial Depth & 1 (Root = 0)\\ 
\hspace{.3cm}Final Depth & 5 \\ \hline
Maximum Length & 800 \\ \hline
Maximum Final Depth & 8\\ \hline
Independent Runs & 30 \\ \hline
Semantic Thresholds &  UBSS = 0.5 \\ \hline
Neighbourhood size & 20 \\ \hline
\end{tabular}
}
\label{tab:parameters}
\end{table}

A common metric used in determining fitness for binary classification problems, as the ones modelled in this work, is to use classification accuracy; where $ACC = \frac{TP + TN}{TP + TN + FP + FN}$. However, with imbalanced data sets, using this accuracy measure will tend to bias towards the majority class as shown in~\cite{6198882}. As such it is better to treat the minority and majority as two separate objectives where the goal is to maximize the number of correctly classified cases. This can be done using the true positive rate $TPR = \frac{TP}{TP + TN}$ and true negative $TNR = \frac{TN}{TN + FP}$  \cite{6198882}.  Table \ref{tab:parameters} gives an overview of the parameters used in our work. These values are the result of preliminary experiments not reported in this work. In particular, the value for the semantic threshold yields promising results. This is in agreement with other studies indicating that this value successfully promotes semantics helping evolutionary search~\cite{Galvan_MICAI_2016, DBLP:conf/gecco/GalvanS19, Uy2011}.

\section{Results}
\label{sec:results}

Table \ref{tab:moead} reports both the average hypervolume over 30 independent runs. We also computed the accumulated Pareto-optimal (PO) front with respect to 30 runs: the set of non-dominated solutions after merging all 30 Pareto-approximated fronts. These results were gathered for a canonical MOEAD/D framework and by combining MOEA/D with the SSC method as outlined by Uy et al. ~\cite{Uy2011}. Both methods used the Tchebycheff approach and only consider the majority and minority as objectives. To obtain a statistically sound conclusion, the Wilcoxon rank-sum test was run with a significance level of $\alpha$= 0.05 on the average hypervolume results. These statistically significant differences are highlighted in boldface. By using SSC with MOEA/D-TCH we find that the hypervolume results are significantly better for every digit. 

Furthermore a payoff table was created to compare the results canonical forms of NSGA-II, SPEA2, MOEA/D-TCH and MOEA/D-TCH SSC (for reference the hypervolume results have been included in Table~\ref{tab:hyperarea:nsgaii:spea2}). 
The payoff table can be read as follows; the strategies of the column index are compared against the strategies of the row index and for each digit that is significantly better for the column strategy counts as one `win' towards the count. For example, MOEA/D-TCH SSC produced 10 `wins' over its canonical from but results were comparatively mixed when compared against the canonical forms of NSGA-II and SPEA2. In this regard, SPEA2 performed better for 5 digits versus MOEA/D-TCH SSC which performed better for 2 and a somewhat more even spread of wins for NSGA-II, which performed better for 3 digits versus MOEA/D-TCH SSC which performed better for 2. Both these methods produced significantly better results over the canonical form of MOEA/D garnering 8 `wins' each in this respect. In general we can say that MOEA/D performs worse on the MNIST data set when compared against NSGA-II and SPEA2 but that when SSC is included the performance is somewhat more comparable.

\begin{figure*}[!htbp]
  \centering
  \begin{tabular}{ccc}
     \scriptsize{MNIST 0} & \scriptsize{MNIST 1} & \scriptsize{MNIST 2} \\
     

     \hspace{-0.82cm}  \includegraphics[width=0.370\textwidth]{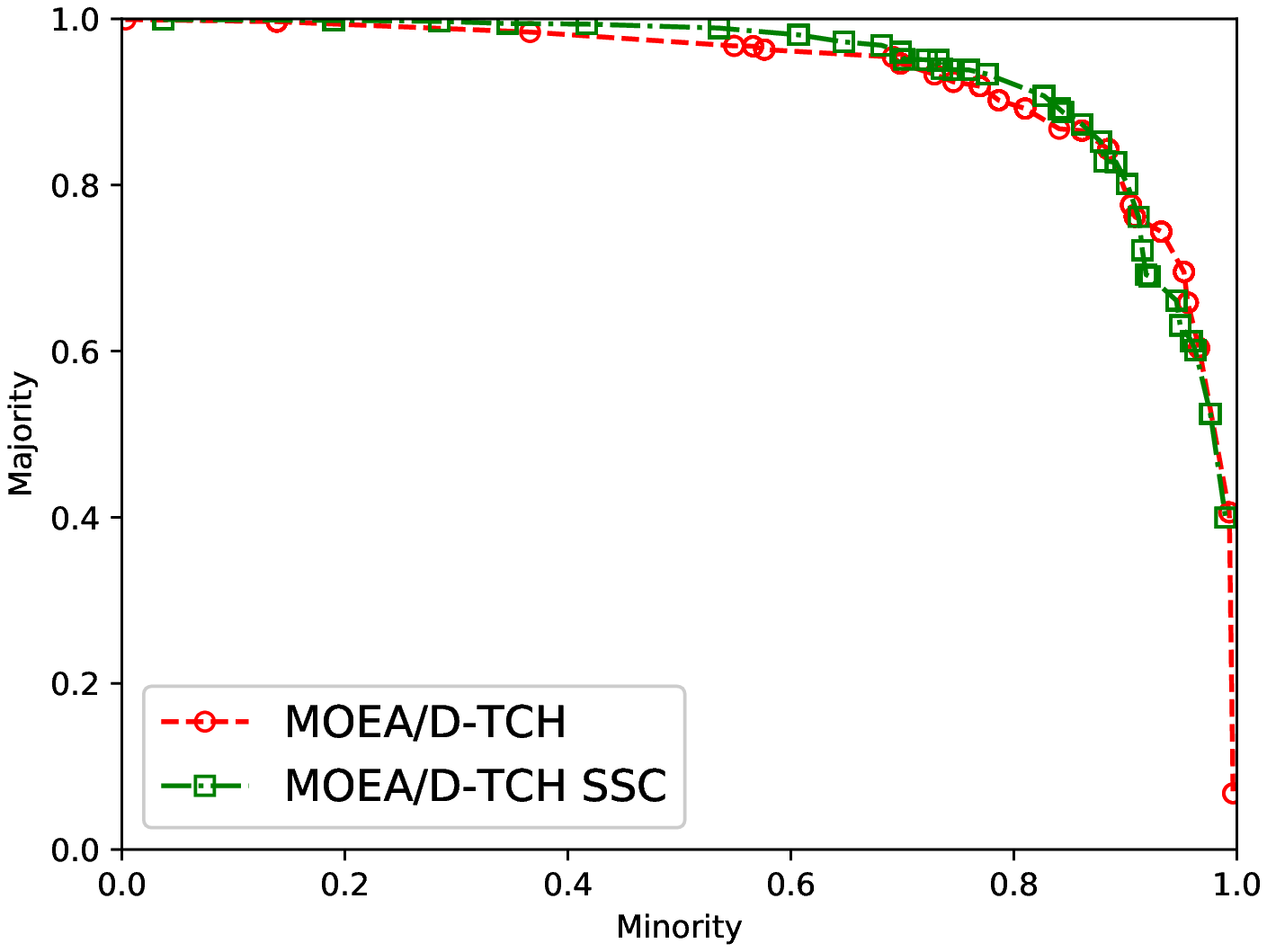}   & \hspace{-0.95cm}  \includegraphics[width=0.370\textwidth]{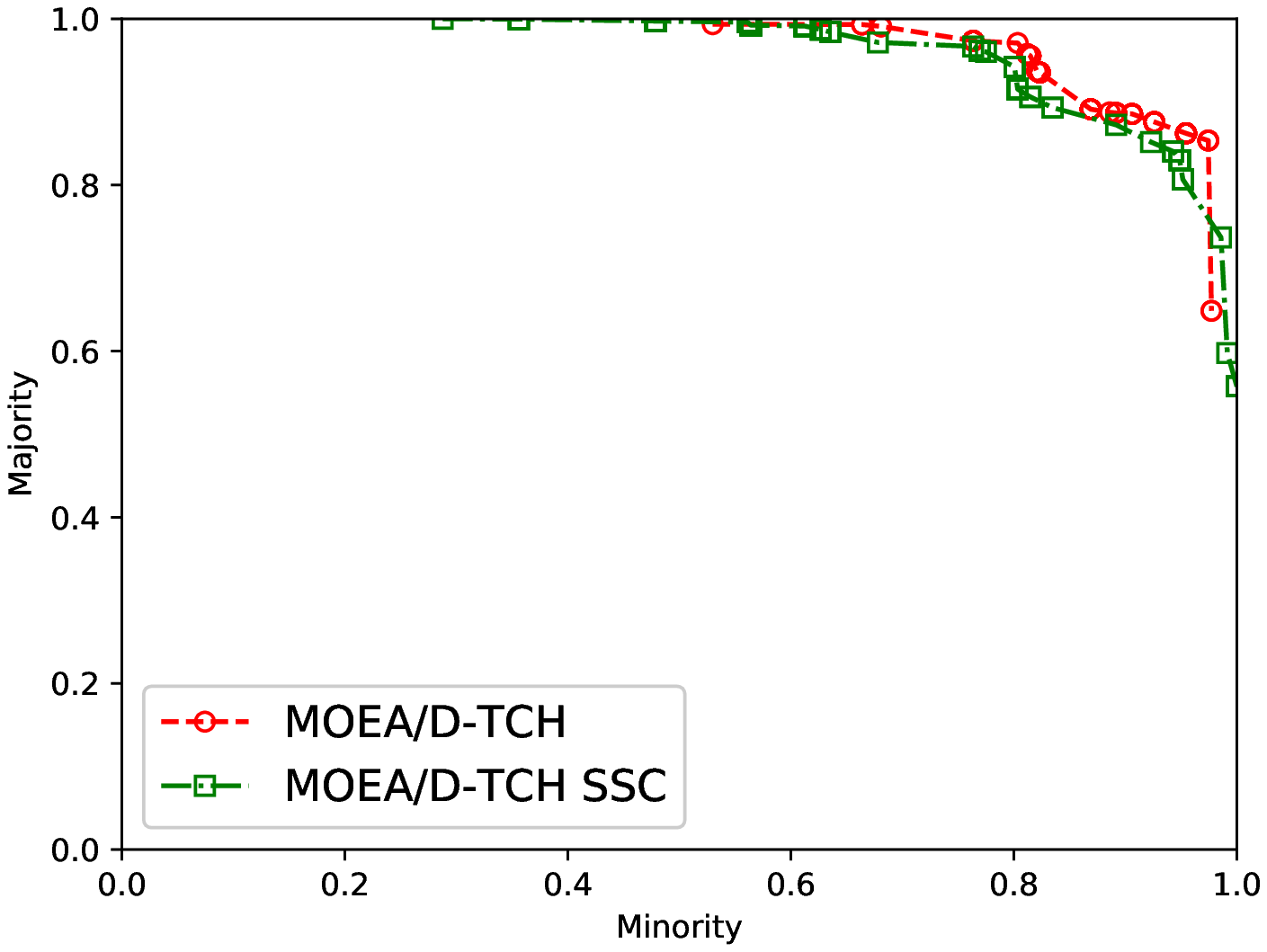} & \hspace{-0.95cm}  \includegraphics[width=0.370\textwidth]{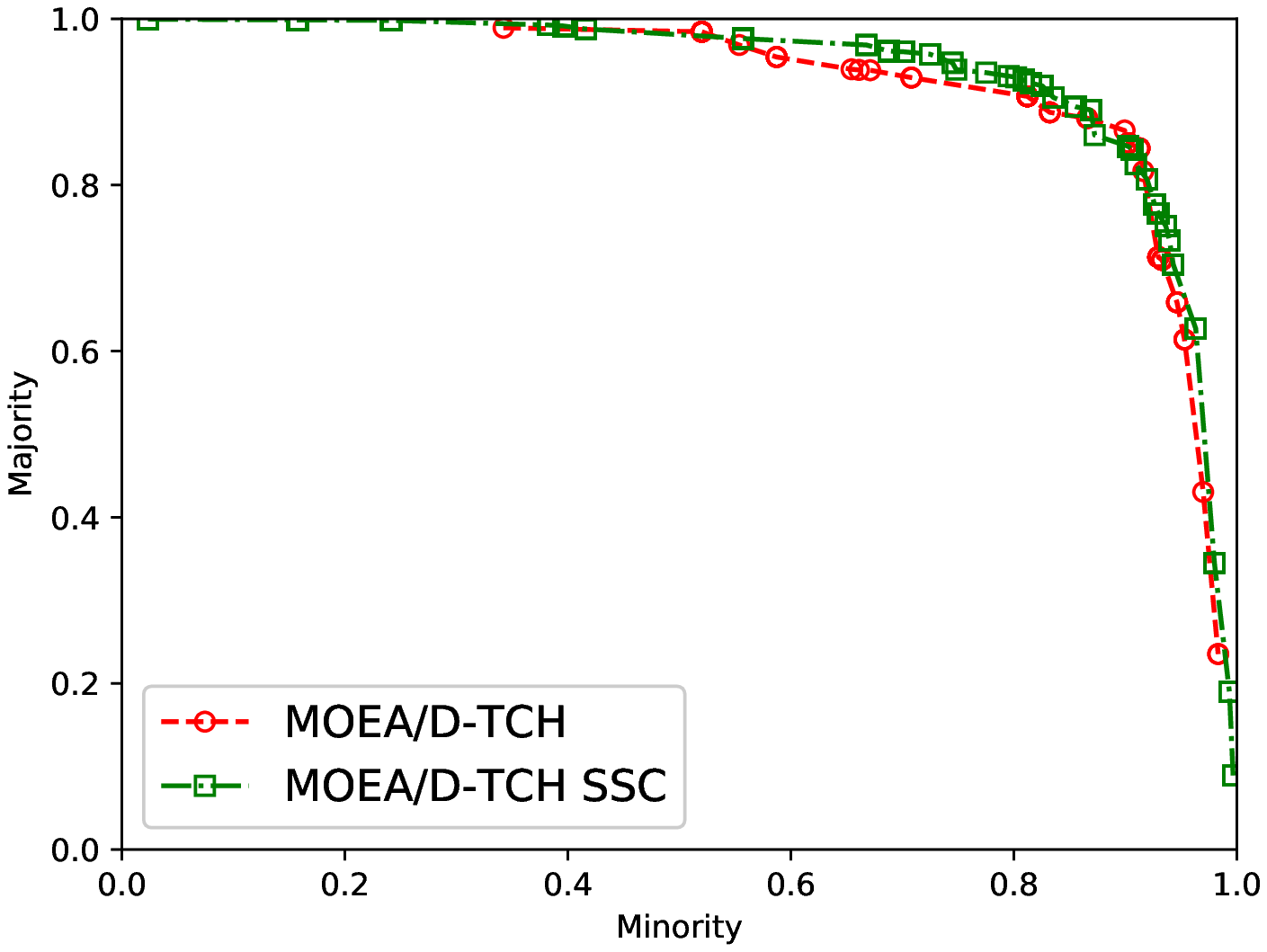} 
\end{tabular}
\begin{tabular}{ccc}
  \scriptsize{MNIST 3} & \scriptsize{MNIST 4} & \scriptsize{MNIST 5} \\


\hspace{-0.82cm}  \includegraphics[width=0.370\textwidth]{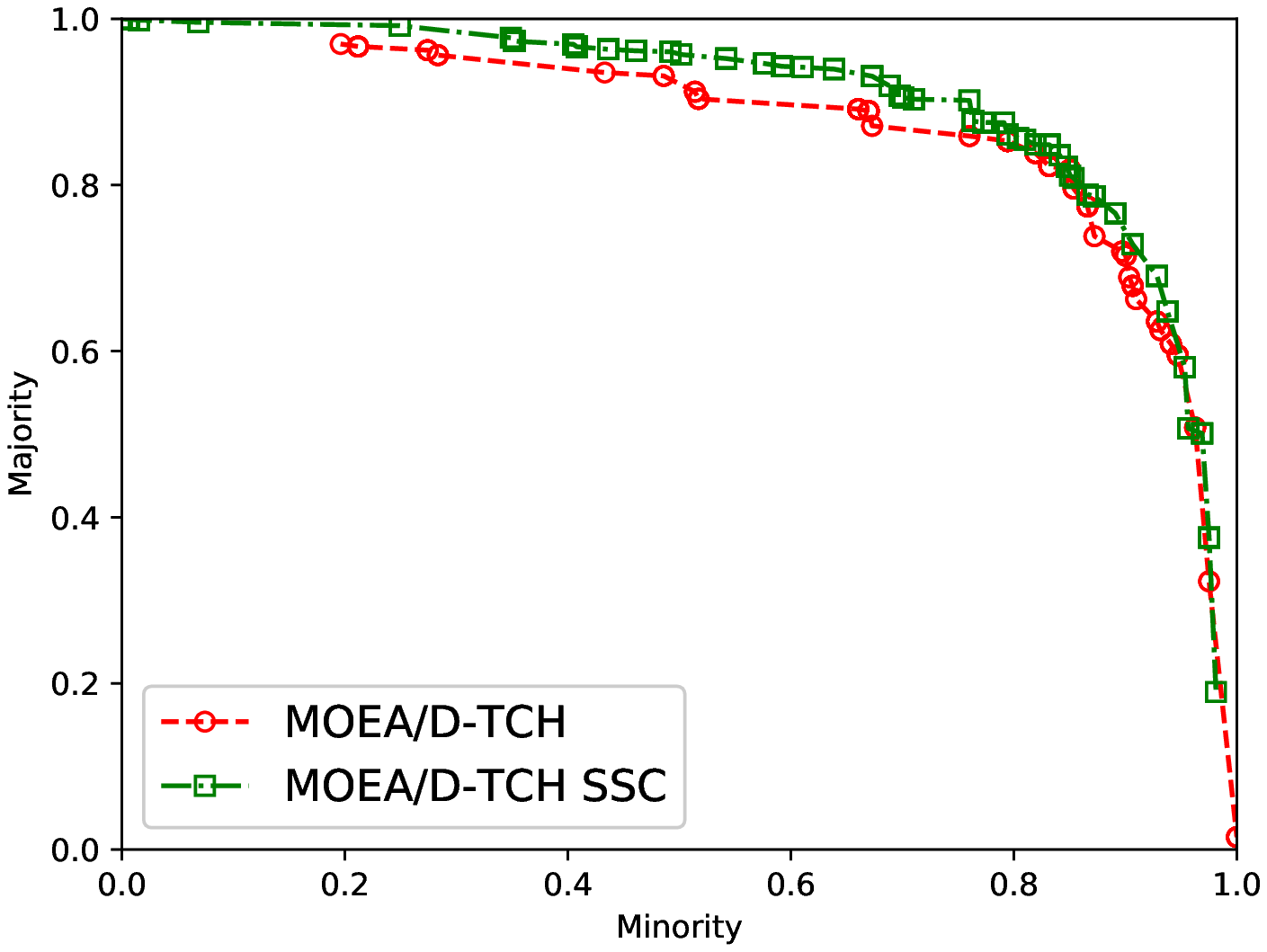}   & \hspace{-0.95cm}  \includegraphics[width=0.370\textwidth]{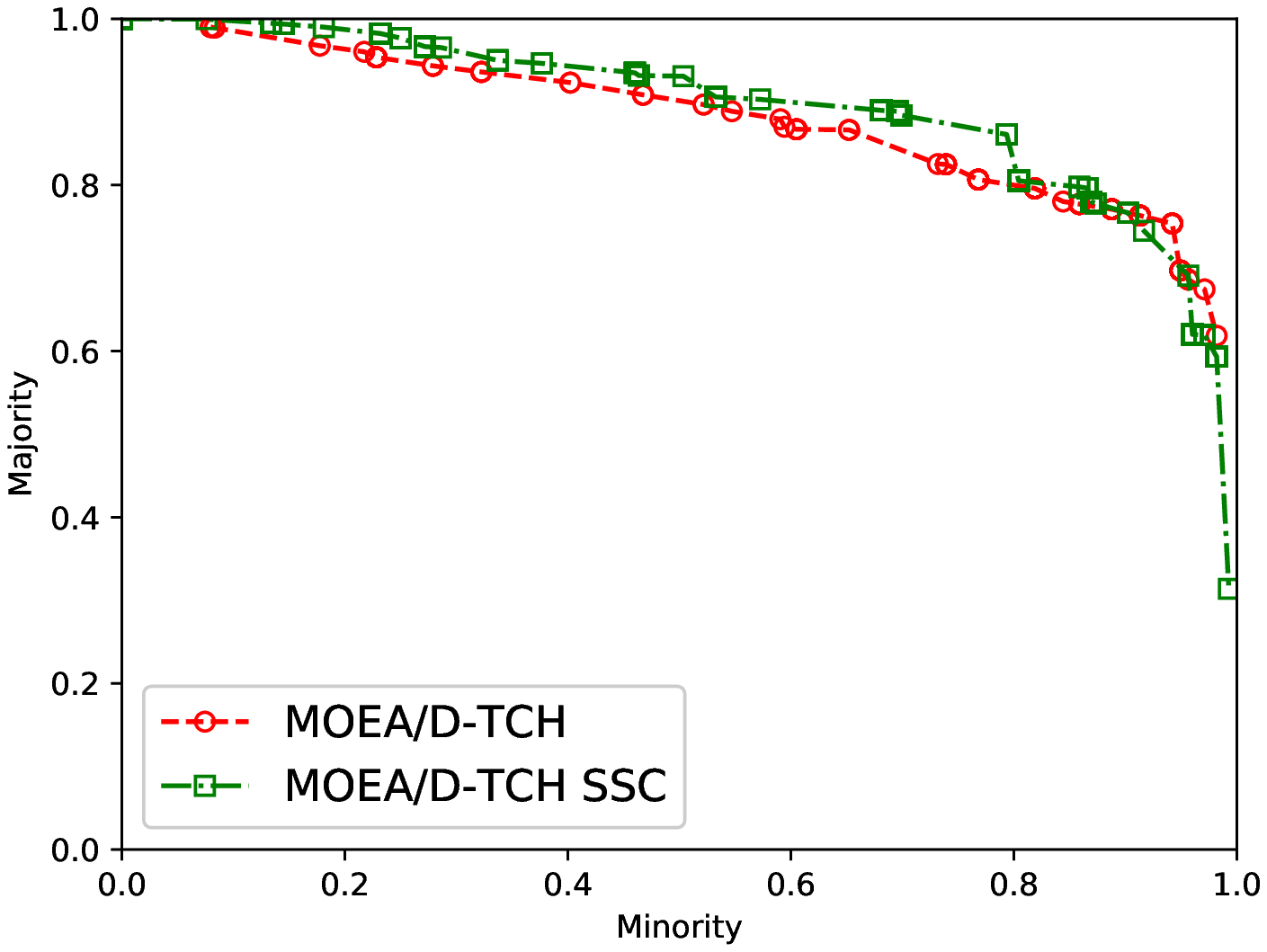} & \hspace{-0.95cm}  \includegraphics[width=0.370\textwidth]{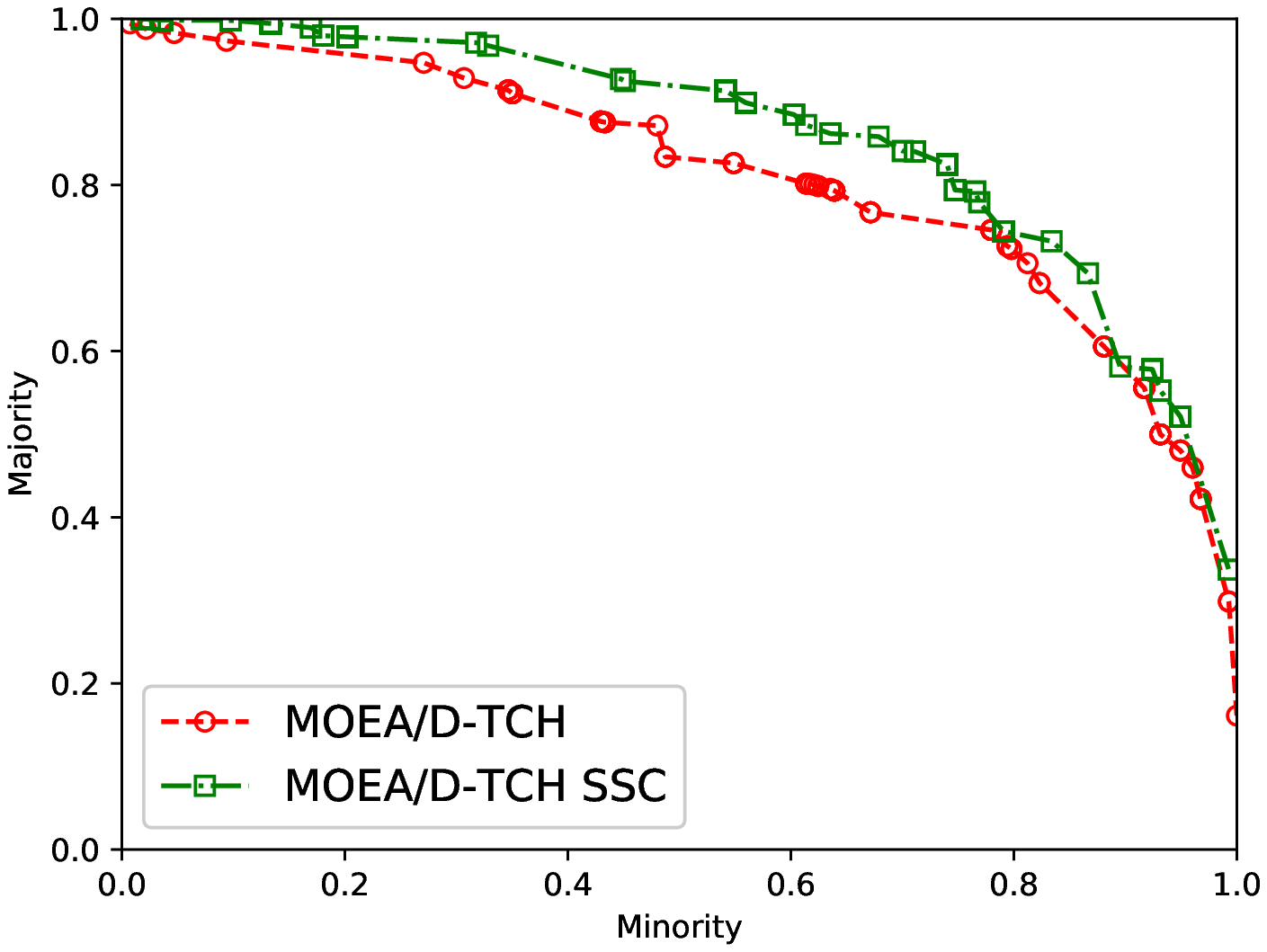} 
\end{tabular}
\begin{tabular}{cc}
	
  \scriptsize{MNIST 6} & \scriptsize{MNIST 7} \\


\hspace{-0.82cm}  \includegraphics[width=0.370\textwidth]{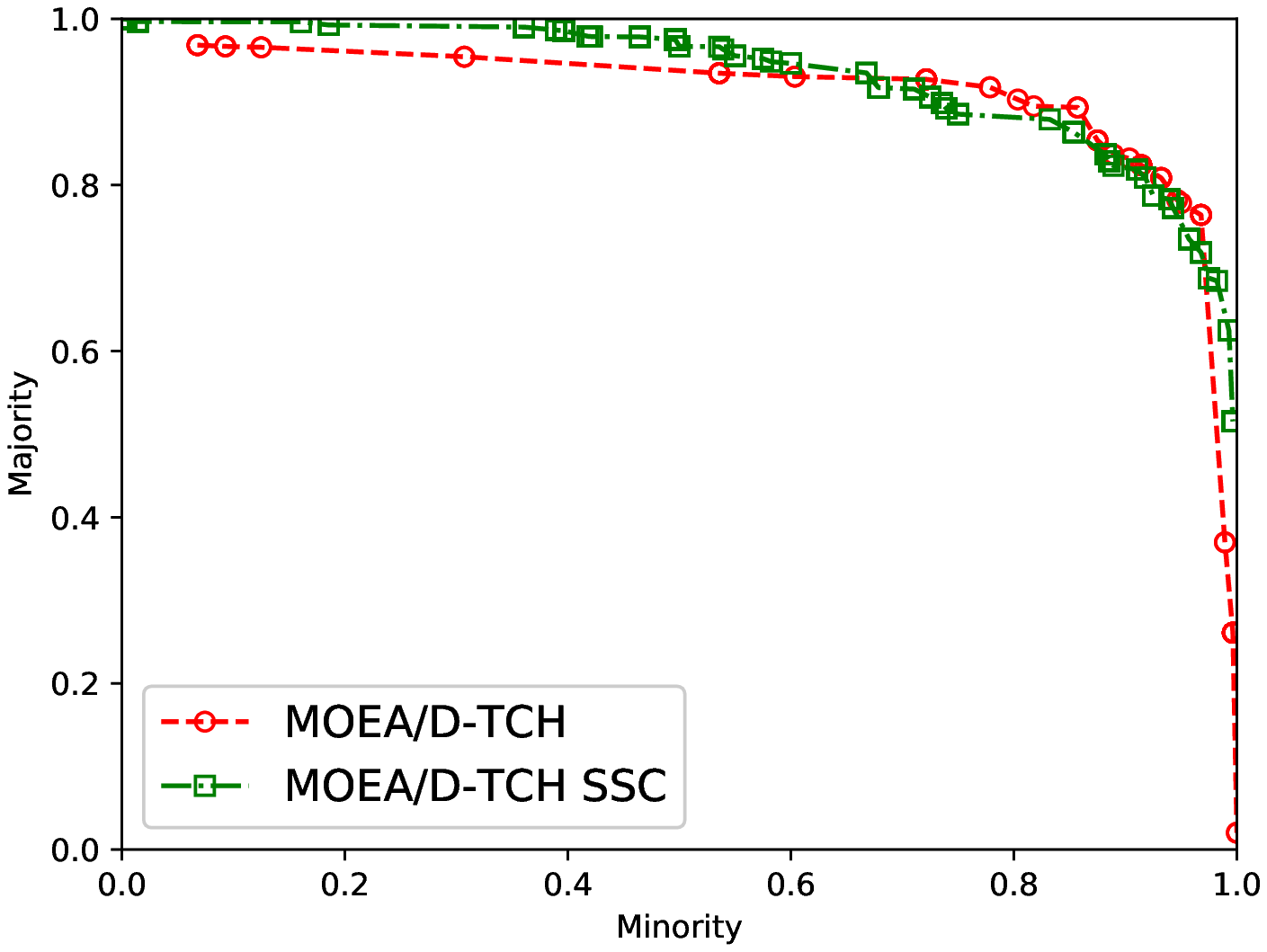}   & \hspace{-0.95cm}  \includegraphics[width=0.370\textwidth]{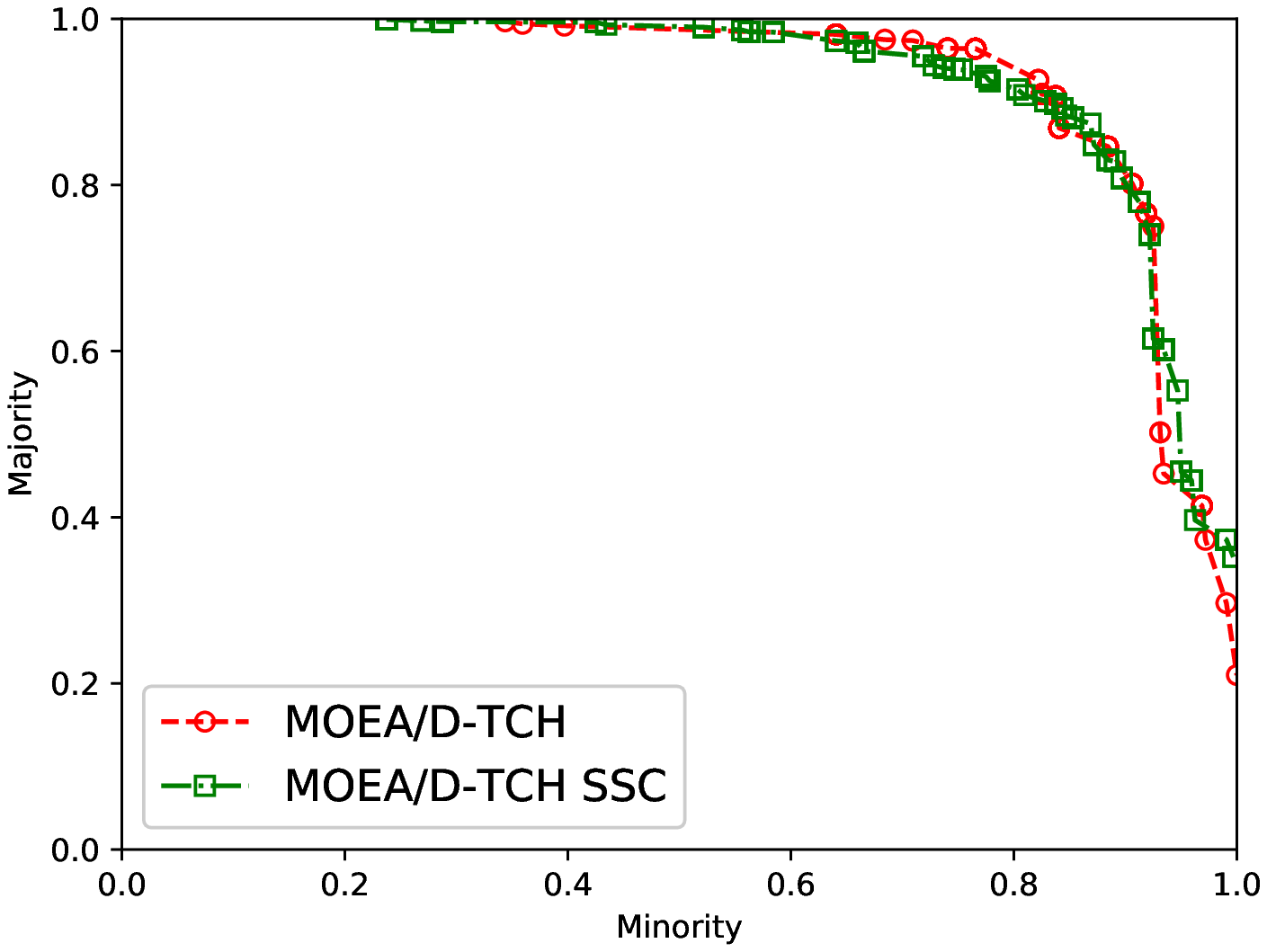}

\end{tabular}
\begin{tabular}{cc}

  \scriptsize{MNIST 8} & \scriptsize{MNIST 9} \\


\hspace{-0.82cm}  \includegraphics[width=0.370\textwidth]{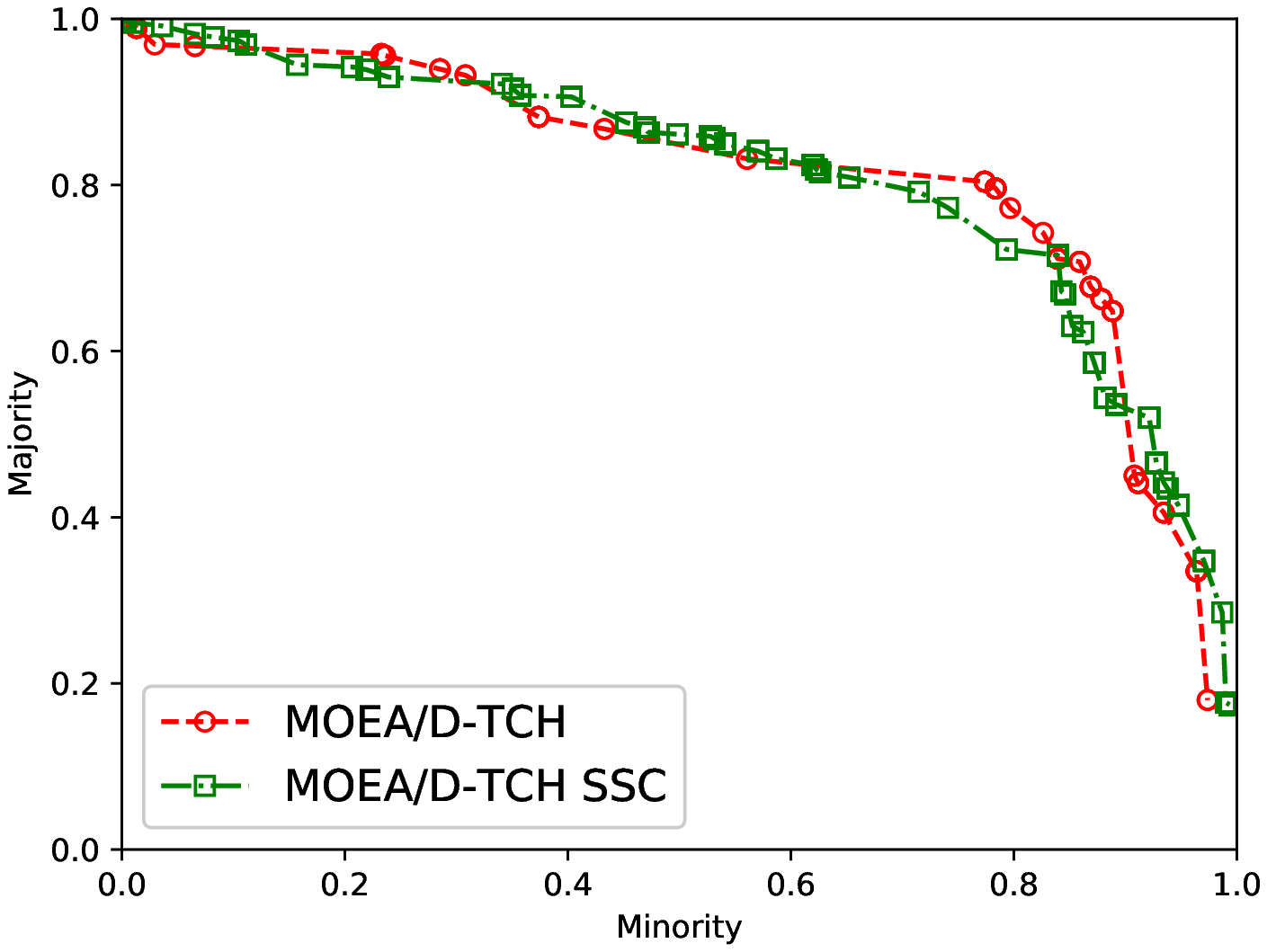}   & \hspace{-0.95cm}  \includegraphics[width=0.370\textwidth]{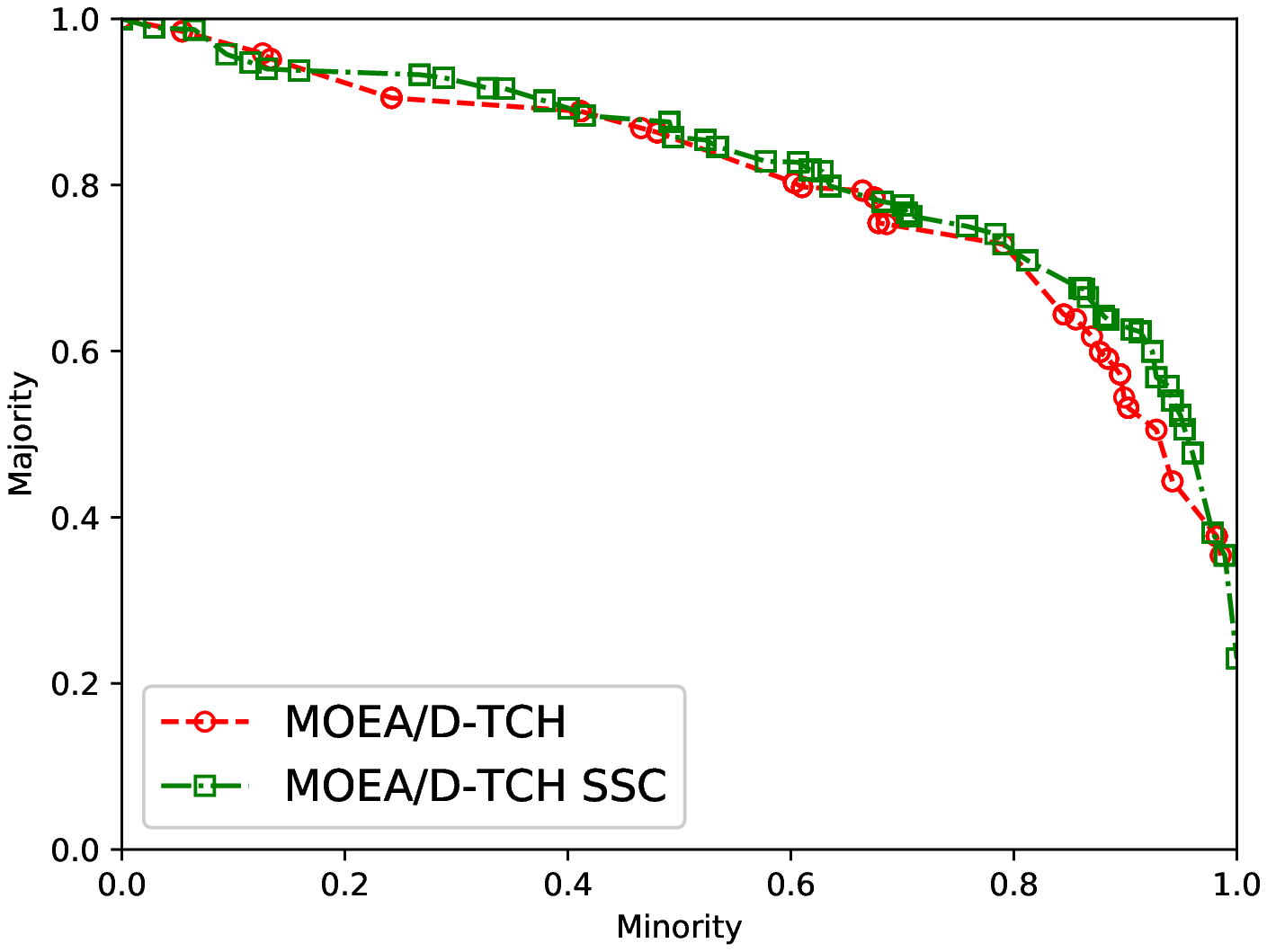}

\end{tabular}
\caption{Pareto fronts for  each of the MNIST digits using MOEA/D-TCH and MOEA/D-TCH SSC.} 
  
\label{fig:pareto}
\end{figure*}

Figure \ref{fig:pareto} shows the accumulated PO with respect to 30 runs for MOEA/D-TCH and MOEA/D-TCH SSC methods for the 10  digits. The SSC method is shown in green and we can see the hypervolume covers a greater area in objective space for the majority and minority classes when compared with the canonical method of MOEA/D-TCH for all 10 digits, although for digits 1 and 8 this distinction is harder to make out.

\begin{table}[tb]
\caption{Average ($\pm$ std deviation) hypervolume of evolved Pareto-approximated fronts and PO fronts  for MOEA/D-TCH and MOEA/D-TCH SSC over 30 independent runs for MNIST data set.}
\centering
\resizebox{1.0\columnwidth}{!}{
\begin{tabular}{ccccc}\hline
\multirow{3}{*}{Dataset}
& 
\multicolumn{2}{c}{MOEA/D-TCH} & \multicolumn{2}{c}{MOEA/D-TCH SSC} \\ & \multicolumn{2}{c}{Hypervolume} & \multicolumn{2}{c}{Hypervolume} \\
    & Average & PO Front 
    & Average & PO Front \\ \hline

Mnist 0 & 0.908 $\pm$ 0.009 & 0.918
& \textbf{0.925 $\pm$ 0.008}+ & 0.928\\

Mnist 1 & 0.945 $\pm$ 0.011 & 0.949 
& \textbf{0.961 $\pm$ 0.005}+ & 0.957\\

Mnist 2 & 0.911 $\pm$ 0.012 & 0.911 
& \textbf{0.925 $\pm$ 0.010}+ & 0.928\\

Mnist 3 & 0.869 $\pm$ 0.020 & 0.860
& \textbf{0.890 $\pm$ 0.014}+ & 0.894\\

Mnist 4 & 0.867 $\pm$ 0.021 & 0.863
& \textbf{0.893 $\pm$ 0.012}+ & 0.886\\

Mnist 5 & 0.822  $\pm$ 0.018 & 0.799
& \textbf{0.850 $\pm$ 0.013}+ & 0.845\\

Mnist 6 & 0.914  $\pm$ 0.012 & 0.903
& \textbf{0.929 $\pm$ 0.009}+ & 0.928\\

Mnist 7 & 0.920 $\pm$ 0.012 & 0.921 
& \textbf{0.931 $\pm$ 0.008}+ & 0.925\\

Mnist 8 & 0.786 $\pm$ 0.021 & 0.795 
& \textbf{0.806 $\pm$ 0.022}+ & 0.803\\

Mnist 9 & 0.783 $\pm$ 0.018 & 0.783 
& \textbf{0.818 $\pm$ 0.018}+ & 0.813 \\

\hline
\end{tabular}
}
\label{tab:moead}
\end{table}

  \begin{table}[tb]
  \centering
    \caption{Payoff tables for canonical NSGA-II, SPEA2, MOEA/D-TCH and the semantic-based method MOEA/D-TCH SSC for each of the 10 digits}
  \resizebox{0.55\columnwidth}{!}{
     \hspace*{-1cm} \begin{tabularx}{0.35\textwidth}{cc|c|c|c|c|}
    
        & \multicolumn{1}{c}{} &

        \multicolumn{1}{Y}{\tiny{A}}  & \multicolumn{1}{Y}{\tiny{B}} & \multicolumn{1}{Y}{\tiny{C}}  & \multicolumn{1}{Y}{\tiny{D}}
      
        \\\cline{3-6}

        \multirow{4}* 
        & \tiny{A} & - & 0 \cellcolor{01}  & \textcolor{white}{8} \cellcolor{89}  &  3 \cellcolor{23} 
      
        \\\cline{3-6}
      
        & \tiny{B} & 0 \cellcolor{01}  & - & \textcolor{white}{8} \cellcolor{89} & \textcolor{white}{5} \cellcolor{45}
        \\\cline{3-6}
              & \tiny{C} & {0} \cellcolor{01} & {1} \cellcolor{01} & - & {0} \cellcolor{01}
        \\
        \cline{3-6}
              & \tiny{D} &2 \cellcolor{23}  & 2 \cellcolor{23}   & \textcolor{white}{10} \cellcolor{10}  & -
        \\\cline{3-6}

      \end{tabularx}}

      \begin{tablenotes}
      \item 
      \item \hspace{70pt}A:\hspace{10pt} NSGA-II       
      \item \hspace{70pt}B:\hspace{10pt} SPEA2         
      \item \hspace{70pt}C:\hspace{10pt} MOEA/D-TCH    
      \item \hspace{70pt}D:\hspace{10pt} MOEA/D-TCH SSC
      \end{tablenotes}

  \end{table}

\section{Conclusions}
\label{sec:conclusions}
To the authors knowledge, this paper presents the first method to promote semantics into an Multi-objective Genetic Programming paradigm using a decomposition method.
While incorporating SSC into SOGP resulted in significantly better results for Uy et al. ~\cite{Uy2011}. Previous works by Galv\'an et al.~\cite{DBLP:conf/gecco/GalvanS19} showed that this did not follow through in a EMO paradigm when compared with canonical forms of NSGA-II and SPEA2. Both these methods use dominance-based fitness criteria to determine selection, where each objective is considered separately. However decomposition uses an aggregated scalarization for fitness, which is more akin to SOGP selection. As such this may explain why incorporating SSC in a MOEA/D framework produces better performance over its canonical form, whereas with the non-decomposition MOGP methods the performance is not significantly better.  This demonstrates an important advantage to using decompositional approaches in MOGP with regards to semantics. 

It was also found that for the purely canonical forms, MOEA/D performed significantly worse than NSGA-II. Future work comparing semantics would benefit by using data sets where MOEA/D is known to perform significantly better than NSGA-II for the MNIST data set. Furthermore only one decomposition approach has been tested for this paper: the Tchebycheff approach. As SSC is performed independently of the decomposition approach used, this method can be tested on a variety of different variants of MOEA/D, in particular PBI and inverted PBI methods have also been reported to perform better than the Tchebycheff approach depending structure of the Pareto front under consideration and as such would be a good starting point for further analysis.



%
%
%
%

\section{Acknowledgments}

\noindent This publication has emanated from research conducted with the financial support of Science Foundation Ireland under Grant number 18/CRT/6049. The authors wish to acknowledge the DJEI/DES/SFI/HEA Irish Centre for High-End Computing (ICHEC) for the provision of computational facilities and support. 


\begin{table}[tb]
\caption{Average ($\pm$ std deviation) hypervolume of evolved Pareto-approximated fronts and PO fronts for canonical NSGA-II and SPEA2 over 30 independent runs for MNIST data set. }
\centering
\resizebox{0.46\textwidth}{!}{
\begin{tabular}{ccccc}\hline
  \multirow{3}{*}{Dataset}
  
& 
\multicolumn{2}{c}{NSGA-II} & \multicolumn{2}{c}{SPEA2} \\ & \multicolumn{2}{c}{Hypervolume} & \multicolumn{2}{c}{Hypervolume} \\
    & Average & PO Front 
    & Average & PO Front \\ \hline

Mnist 0 & 0.927 $\pm$ 0.012 & 0.922 &  {0.926 $\pm$ 0.017} & 0.940  \\

Mnist 1 & 0.963 $\pm$ 0.010 & 0.963 & {0.965 $\pm$ 0.008} & 0.970  \\

Mnist 2 & 0.932 $\pm$ 0.012 & 0.945 & 
0.932 $\pm$ 0.014 & 0.937 \\

Mnist 3 & 0.897 $\pm$ 0.041 & 0.917 & 
0.907 $\pm$ 0.017 & 0.913 \\

Mnist 4 & 0.907 $\pm$ 0.039 & 0.929 & 
0.914 $\pm$ 0.022 & 0.925 \\

Mnist 5 & 0.855 $\pm$ 0.037 & 0.872 & {0.849 $\pm$ 0.022} & 0.877  \\

Mnist 6 & 0.931 $\pm$ 0.016 & 0.947 & {0.929 $\pm$ 0.019} & 0.944 \\

Mnist 7 & 0.939 $\pm$ 0.007 & 0.947 & {0.937 $\pm$ 0.008} & 0.946 \\

Mnist 8 & 0.770 $\pm$ 0.094 & 0.819 & 
0.770 $\pm$ 0.084 & 0.756 \\

Mnist 9 & 0.758 $\pm$ 0.097 & 0.831 & {0.747 $\pm$ 0.085} & 0.712 \\

\hline
\end{tabular}
}
\label{tab:hyperarea:nsgaii:spea2}
\end{table}

\bibliographystyle{abbrv}
\bibliography{moed_galvan_v2.bib}

\end{document}